\documentclass[conference]{IEEEtran}
\IEEEoverridecommandlockouts
\usepackage{cite}
\usepackage{amsmath,amssymb,amsfonts}
\usepackage{algorithmic}
\usepackage{graphicx}
\usepackage{textcomp}
\usepackage{xcolor}
\usepackage[raggedright]{sidecap}
\usepackage[singlelinecheck=false,justification=justified]{caption}
\usepackage{booktabs}

\usepackage{sidecap}
\sidecaptionvpos{figure}{t}

\def\BibTeX{{\rm B\kern-.05em{\sc i\kern-.025em b}\kern-.08em
    T\kern-.1667em\lower.7ex\hbox{E}\kern-.125emX}}
\begin{document}

\title{Fusion and Orthogonal Projection for Improved Face-Voice Association}

\author{Muhammad Saad Saeed$^{1}$, Muhammad Haris Khan$^{2}$, Shah Nawaz$^{3}$, \\ Muhammad Haroon Yousaf$^{1}$, Alessio {Del Bue}$^{3,4}$  \\
$^{1}$University of Engineering and Technology Taxila,
$^{2}$Muhammad Bin Zayed University of Artificial Intelligence,\\ 
$^{3}$Pattern Analysis \& Computer Vision (PAVIS) - Istituto Italiano di Tecnologia (IIT),\\ 
$^{4}$Visual Geometry \& Modelling (VGM) - Istituto Italiano di Tecnologia (IIT)}


\maketitle

\begin{abstract}
We study the problem of learning association between face and voice, which is gaining interest in the computer vision community lately.
Prior works adopt pairwise or triplet loss formulations to learn an embedding space amenable for associated matching and verification tasks.  
Albeit showing some progress, such loss formulations are, however, restrictive due to dependency on distance-dependent margin parameter, poor run-time training complexity, and reliance on carefully crafted negative mining procedures.
In this work,  we hypothesize that enriched feature representation coupled with an effective yet efficient supervision is necessary in realizing a discriminative joint embedding space for improved face-voice association.
To this end, we propose a light-weight, plug-and-play mechanism that exploits the complementary cues in both modalities to form enriched fused embeddings and clusters them based on their identity labels via orthogonality constraints.
We coin our proposed mechanism as fusion and orthogonal projection (FOP) and instantiate in a two-stream pipeline. The overall resulting framework is evaluated on a large-scale VoxCeleb dataset with a multitude of tasks, including cross-modal verification and matching. 
Results show that our method performs favourably against the current state-of-the-art methods and our proposed supervision formulation is more effective and efficient than the ones employed by the contemporary methods.
\end{abstract}


\section{Introduction}
It is a well-studied and understood fact that humans can associate voices and faces of people because the neuro-cognitive pathways for voices and faces share same structure~\cite{kamachi2003putting,belin2004thinking}. Recently, Nagrani et al.~\cite{nagrani2018seeing,nagrani2018learnable,nagrani2017voxceleb} introduced the face-voice association task into vision community with the creation of a large-scale audio-visual dataset, comprising faces and voices of $1,251$ celebrities. Since then, the face-voice association task has gained significant research interest~\cite{horiguchi2018face,kim2018learning,nagrani2018learnable,nagrani2018seeing,nawaz2019deep,wen2021seeking,wen2018disjoint}. In addition, we are also witnessing creation of new audio-visual datasets to study this novel task. For example, Nawaz et al.~\cite{nawaz2021cross} introduced a Multilingual Audio-Visual (\textit{MAV-Celeb}) dataset to analyze the impact of language on face-voice association task; it comprises of video and audio recordings of different celebrities speaking more than one language.

Most existing works ~\cite{kim2018learning,nagrani2018seeing,nagrani2018learnable,nawaz2021cross} tackle face-voice association as a cross-modal biometric task. The two prominent challenges in developing an effective method for this task are learning of a common yet discriminative embedding space, where instances from two modalities are sufficiently aligned and instances of semantically similar identities are nearby. Often separate networks for face and voice modalities are leveraged to obtain the respective feature embeddings and contrastive or triplet loss formulations are employed to construct this embedding space. 
Although showing some effectiveness in this task, such loss formulations, however, are restrictive in following ways. First, they require tuning of a margin hyperparameter, which is hard as the distances between instances can alter significantly while training. Secondly, the run-time training complexity for contrastive and triplet losses are $\mathcal{O} (n^2)$ and $\mathcal{O} (n^3)$, respectively, where $n$ is the number of available instances for a modality. Finally, to mitigate the high run-time training complexity challenge, different variants of carefully crafted negative mining strategies are used, which are both time-consuming and performance sensitive. 

A few methods e.g., \cite{nawaz2019deep} have attempted to replace the contrastive/triplet loss formulations by utilizing auxiliary identity centroids~\cite{wen2016discriminative}. The training process alternates between the following two steps: 1) clustering embeddings around their identity centroids and pushing embeddings away from all other identity centroids, and 2) updating these centroids using the mini-batch instances. Such centroid based losses are used with traditional classification loss (i.e. softmax cross-entropy (CE)). However, their co-existence is unintuitive and ineffective because the former promotes margins in Euclidean space whereas latter implicitly achieves separability in the angular domain.

In this work, we hypothesize that an enriched unified feature representation, encompassing complementary cues from both modalities, alongside an effective yet efficient supervision formulation is crucial towards realizing a discriminative joint embedding space for improved face-voice association. To this end, we propose a light-weight, plug-and-play mechanism that exploits the best in both modalities through fusion and semantically aligns fused embeddings with their identity labels via orthogonality constraints. We instantiate our proposed mechanism in the two-stream pipeline, which provides face and voice embeddings, and the resulting overall framework is an effective and efficient approach for face-voice matching and verification tasks.

We summarize our key contributions as follows. 1) We propose to harness the complementary features from both modalities in forming enriched feature embeddings, that are consistent with semantics of identity, thereby allowing improved identity recognition. 2) We propose to impose orthogonality constraints on the fused embeddings. They are not only coherent with the angular characteristic of the commonly employed classification loss but are very efficient as they operate directly on mini-batches. 3) Experimental results on large-scale VoxCeleb \cite{nagrani2017voxceleb} show the effectiveness of our method on both face-voice verification and matching tasks. Further, we note that our method performs favourably against the existing state-of-the-art methods. 4) We perform a thorough ablation study to analyze the impact of different components. 

\section{Related Work}


\noindent \textbf{Face-voice Association.}
The work of Nagrani et al.~\cite{nagrani2018seeing} leveraged audio and visual information to establish an association between faces and voices in a cross-modal biometric matching task.
Similarly, some recent work~\cite{kim2018learning,nagrani2018learnable,wen2021seeking,nawaz2021cross} introduced joint embeddings to establish correlation between face and voice of an identity. These methods extract audio and face embeddings and then minimize the distance between embeddings of same identities while maximize the distance among embeddings from different ones.
Wen et al.~\cite{wen2018disjoint} presented a disjoint mapping network to learn a shared representation for audio and visual information by mapping them individually to common covariates (gender, nationality, identity) without needing to construct pairs or triplets at the input. 
Similarly, Nawaz et al.~\cite{nawaz2019deep} extracted audio and visual information with a single stream network to learn a shared deep latent representation, leveraging identity centroids to eliminate the need of pairs or triplets~\cite{nagrani2018learnable,nagrani2018seeing}. 
Both Wen et al.~\cite{wen2018disjoint} and Nawaz et al.~\cite{nawaz2019deep} show that effective face-voice representations can be learned without pairs or triplets formation.


Contrary to previous works, our method proposes to construct enriched embeddings via exploiting complementary cues from the embeddings of both modalities through a attention-based fusion. Further, it clusters the embeddings of same identity and separates embeddings of different identities via orthogonality constraints. The instantiation of both proposals in a two-stream pipeline results in an effective and efficient face-voice association framework.

\begin{figure*}

\includegraphics[scale=0.54]{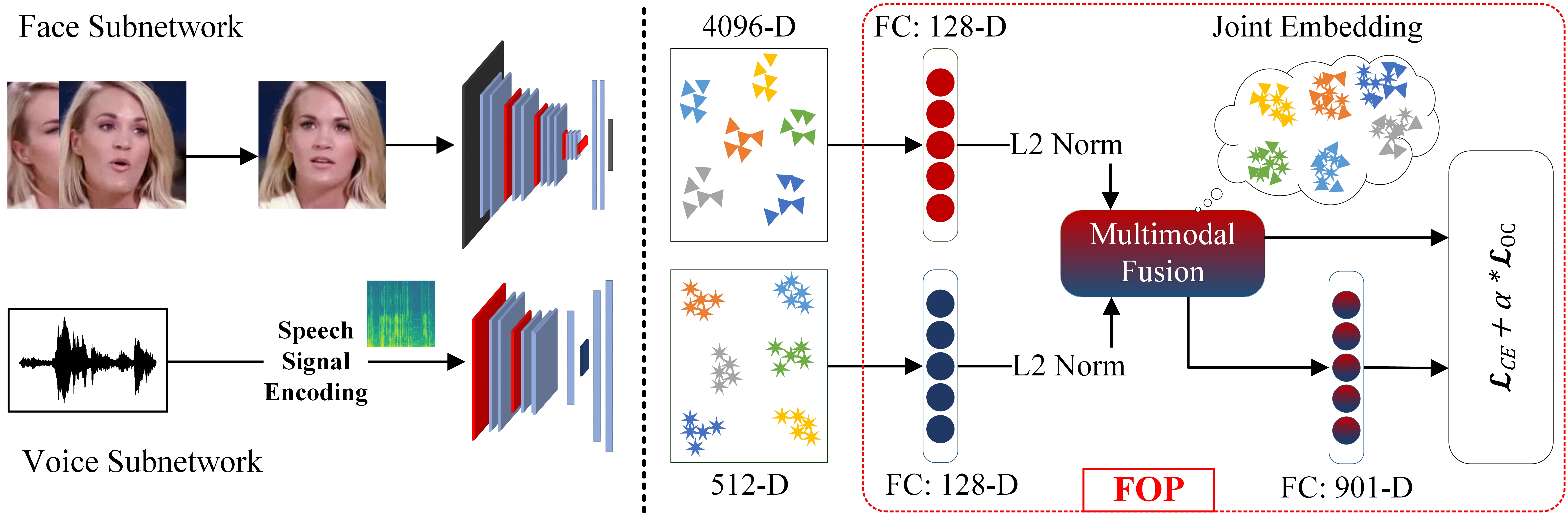}
\includegraphics[scale=0.7]{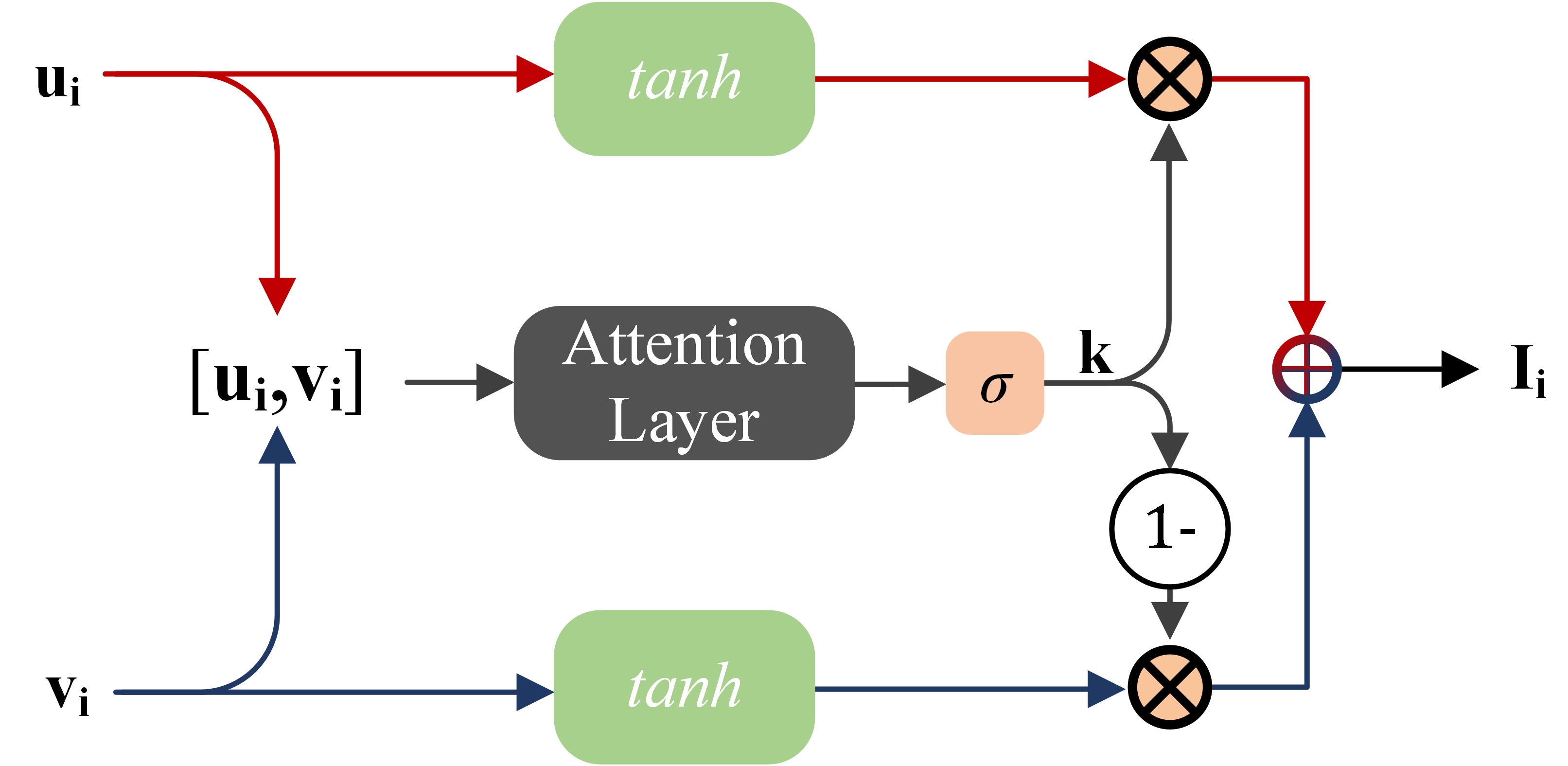}

   \caption{\small (Left) Overall architecture of our method. Fundamentally, it is a two-stream pipeline which generates face and voice embeddings. We propose a light-weight, plug-and-play mechanism, dubbed as fusion and orthogonal projection (FOP) (shown in dotted red box).(Right) The architecture of multimodal fusion. 
   } \vspace{-1em}
\label{fig:overall_framework_fusion}
\vspace{-1em}
\end{figure*}

\section{Overall Framework}
\label{section:overall_framework}

To learn a discriminative joint face-voice embedding for F-V association tasks, we develop a new framework for cross-modal face-voice association (See Fig.~\ref{fig:overall_framework_fusion}) that is fundamentally a two-stream pipeline (sec.~\ref{subsection:Preliminaries}) and features a light-weight module that exploits complementary cues from both face and voice embeddings and facilitates discriminative identity mapping via orthogonality constraints (sec.~\ref{subsection:Learning_Discriminative_Joint_Embeddings}).

\subsection{Preliminaries}
\label{subsection:Preliminaries}
\noindent \textbf{Problem Settings.}
Without the loss of generality, we consider cross-modal retrieval of bimodal data, i.e., for face and voice. Given that we have N instances of face-voice pairs, $\mathcal{D}=\{(x_{i}^{f},x_{i}^{v})\}_{i=1}^{N}$, where $x_{i}^{f}$ and $x_{i}^{v}$ are the face and voice examples of the $i_{th}$ instance, respectively. Each pair of an instance $(x_{i}^{f},x_{i}^{v})$  has an associated label $y_{i}\in\{0,1\}$, where $y_{i}=1$ if $x_{i}^{f}$ and $x_{i}^{v}$ belong to the same identity and $y_{i}=0$ if $x_{i}^{f}$ and $x_{i}^{v}$ belong to a different identity.
Both face and voice embeddings typically lie in different representation spaces owing to their different superficial statistics and are mostly unaligned semantically, rendering them incomparable for cross-modal tasks. Cross-modal learning aims at projecting both into a common yet discriminative representation space, where they are sufficiently aligned and instances from the same identity are nearby while from a different identity are far apart.

\noindent \textbf{Two-stream pipeline.} We employ a two-stream pipeline \cite{nagrani2018learnable} to obtain the respective feature embeddings of both face and voice inputs. The first stream corresponds to a pre-trained convolutional neural network (CNN) on image modality. We take the penultimate layer's output, denoted as $\mathbf{b}_i$, of this CNN as the feature embeddings for an input face image. Likewise, the second stream is a pre-trained audio encoding network that outputs a feature embedding, denoted as $\mathbf{e}_{i}$, for an input audio signal (typically a short-term spectrogram). Existing approaches handling face-voice retrieval \cite{nagrani2018learnable,nawaz2019deep}, mostly resort to triplet and contrastive objectives with carefully crafted negative mining strategies, which significantly increases computational time and are performance-sensitive, to learn a discrminative embedding space. To this end, we introduce a light-weight mechanism that exploits complementary cues from both modality embeddings to form enriched fused embeddings and imposes orthogonal constraints on them for learning discriminative joint face-voice embeddings.

\subsection{Learning Discriminative Joint Embedding}
\label{subsection:Learning_Discriminative_Joint_Embeddings}

In this section, we first describe extracting complementary cues, via multimodal fusion, from both face and voice embeddings obtained through their respective pre-trained networks. We then discuss clustering fused embeddings belonging to the same identity and pushing away the ones with different identity via orthogonality constraints.

Prior to multimodality fusion, we project the face embeddings $\mathbf{b}_i \in \mathbb{R}^{F}$ to a new d-dimensional embedding space $ \mathbf{u}_i \in \mathbb{R}^{d}$ with a fully-connected layer. Similarly, we project the voice embedding $\mathbf{e}_i \in \mathbb{R}^{V}$ to a similar d-dimensional embedding space $ \mathbf{v}_{i} \in \mathbb{R}^{d}$ with another fully-connected layer. We then L2 normalize both $\mathbf{u}_i$ and $\mathbf{v}_i$ which can now be fused to get $\mathbf{l}_{i}$, using the procedure described next.

\noindent \textbf{Multimodal fusion.} We propose to extract complementary features from both modalities, some of which could be related to age, gender and nationality, to form an enriched unified feature representation which is crucial towards learning a discriminative joint embedding space.
Inspired by \cite{arevalo2017gated, chen2020multi}, we employ an attention mechanism to first compute the attention scores (affinity) between the embeddings of two modalities and then fuse these individual modality embeddings after recalibrating them with the attention scores (see Fig.~\ref{fig:overall_framework_fusion}). 
%
We compute attention scores $\mathbf{k}$ between $\mathbf{u}_i$ and $\mathbf{v}_i$  as:
\begin{equation}
    \mathbf{k} = \sigma(F_{att}([\mathbf{u}_i,\mathbf{v}_i]),
\end{equation}

\noindent where $\sigma$ is a sigmoid operator, and $F_{att}$ are the attention layers. Finally, we fuse $\mathbf{u}_i$ and $\mathbf{v}_i$ after modulating them with the attention scores $\mathbf{k}$ to obtain the fused embeddings $\mathbf{l}_{i}$ as:

\begin{equation}
   \mathbf{l}_{i}  = \mathbf{k} \odot tanh(\mathbf{u}_i) + (1 - \mathbf{k}) \odot tanh(\mathbf{v}_i),
\end{equation}

\noindent where $\odot$ is element-wise multiplication.


\noindent \textbf{Supervision via orthogonality constraints.} We want the fused embeddings to encapsulate the semantics of the identity. In other words, these embeddings should be able to predict the identity labels with good accuracy. This is possible if the instances belonging to the same identity are placed nearby whereas the ones with different identity labels are far away. A popular choice to achieve this is softmax cross entropy (CE) loss, which also allows stable and efficient training. Specifically, we use an identity linear classifier with weights denoted as $\mathbf{W}=[\mathbf{w}_{1},\mathbf{w}_{2},...,\mathbf{w}_{C}] \in \mathbb{R}^{d \times C}$ to compute the logits corresponding to $\mathbf{l}_{i}$. Where $d$ is the dimensionality of embeddings and $C$ is the number of identities. Now, identity classification loss with fused embeddings is computed as:

\begin{equation}
    \mathcal{L}_{CE} = -log \frac{exp(\mathbf{l}_{i}^{T}\mathbf{w}_{y_{i}})}{\sum_{j=1}^{C}exp(\mathbf{l}_{i}^{T}\mathbf{w}_{j})}
\end{equation}

Since softmax CE loss does not enforce margins between pair of identities, it is prone to constructing differently-sized class regions which affects identity separability \cite{deng2019arcface,hayat2019gaussian}. 
Some works attempt to include margin between classes in the Euclidean space \cite{wen2016discriminative,calefati2018git}, which is not well synergized with the CE loss as it achieves separation in the angular domain.
Therefore, we propose to impose orthogonality constraints on the fused embeddings to explicitly minimize intra-identity separation while maximizing inter-identity separability \cite{ranasinghe2021orthogonal}. These constraints complement better with the innate angular characteristic of CE loss. Further, since they directly operate on mini-batches, they show greater training efficiency compared to the complex negative mining procedures required in contrastive and triplet loss formulations \cite{nagrani2018learnable, schroff2015facenet} (sec.~\ref{section:experi}).
Formally, the constraints enforce fused embeddings of different identities to be orthogonal and the fused embeddings with same identity to be similar:

\begin{equation}
  \mathcal{L}_{OC}  =  1 - \sum_{i,j \in B, y_{i}=y_{j}} \langle \mathbf{l}_{i},\mathbf{l}_{j}\rangle + \displaystyle\left\lvert \sum_{i,j \in B, y_{i} \neq y_{k}} \langle \mathbf{l}_{i},\mathbf{l}_{k}\rangle\right\rvert,
    \label{Eq:OC}
\end{equation}

\noindent where $ \langle.,.\rangle$ is the cosine similarity operator, and $B$ represents the mini-batch size. The first term in Eq.~\ref{Eq:OC} ensures intra-identity compactness, while the second term enforces inter-identity separation. Note that, the cosine similarity involves the normalization of fused embeddings, thereby projecting them to a unit hyper-sphere:

\begin{equation}
    \langle \mathbf{l}_{i},\mathbf{l}_{j}\rangle = \frac{\mathbf{l}_{i}.\mathbf{l}_{j}}{\lVert \mathbf{l}_{i} \rVert_{2}. \lVert \mathbf{l}_{j} \rVert_{2}}.
\end{equation}

\noindent \textbf{Overall Training Objective.} To train the proposed framework, we minimize the joint loss formulation, comprising of $\mathcal{L}_{CE}$ and $\mathcal{L}_{OC}$ as:


\begin{equation}
\label{eq:floss}
    \mathcal{L} = \mathcal{L}_{CE} + \alpha \mathcal{L}_{OC},
\end{equation}

\noindent where $\alpha$ balances the contribution of two terms in $\mathcal{L}$. We empirically set $\alpha$ to 1.0 based on validation set performance. It is important to mention that CE loss operates in logit space and orthogonal constraints are imposed in the embedding space, however, both of them synergizes well with each other owing to their common angular domain characteristic.

\section{Experiments}
\label{section:experi}


\noindent \textbf{Training Details and Dataset.} We train our method on Quadro P5000 GPU for $50$ epochs using a batch-size of $128$ using Adam optimizer with exponentially decaying learning rate (initialised to \(10^{-5}\)). We extract face and voice embeddings from VGGFace~\cite{parkhi2015deep} and Utterance Level Aggregation~\cite{xie2019utterance}. Note that, we only backprop. through FOP module while the weights of face and voice subnetworks remains unaltered. We perform experiments on \textit{cross-modal verification} and \textit{cross-modal matching} tasks on the large-scale dataset of audio-visual human speech videos ~\cite{nagrani2017voxceleb}. We follow the same train, validation and test split configurations as used in~\cite{nagrani2018learnable} to evaluate on \textit{seen-heard} and \textit{unseen-unheard} identities.

\subsection{Results}
\noindent \textbf{Comparison with other F-V losses.} We compare our (joint) loss formulation against various losses typically employed in F-V association methods, including \textit{center loss}~\cite{wen2016discriminative, nawaz2019deep}, \textit{Git loss} ~\cite{calefati2018git}, \textit{Contrastive Loss} ~\cite{nagrani2018learnable}, and \textit{Triplet Loss} ~\cite{nagrani2018seeing}.
%
%
%
Table~\ref{tab:result-base} reveals that our proposal performs better than other loss formulations across all configurations and both error metrics. Likewise, Table~\ref{tab:runtime-result-base} shows that our (joint) loss formulation is superior than others in terms of both theoretical and empirical training efficiency.

\begin{figure}

    \includegraphics[width=0.24\textwidth]{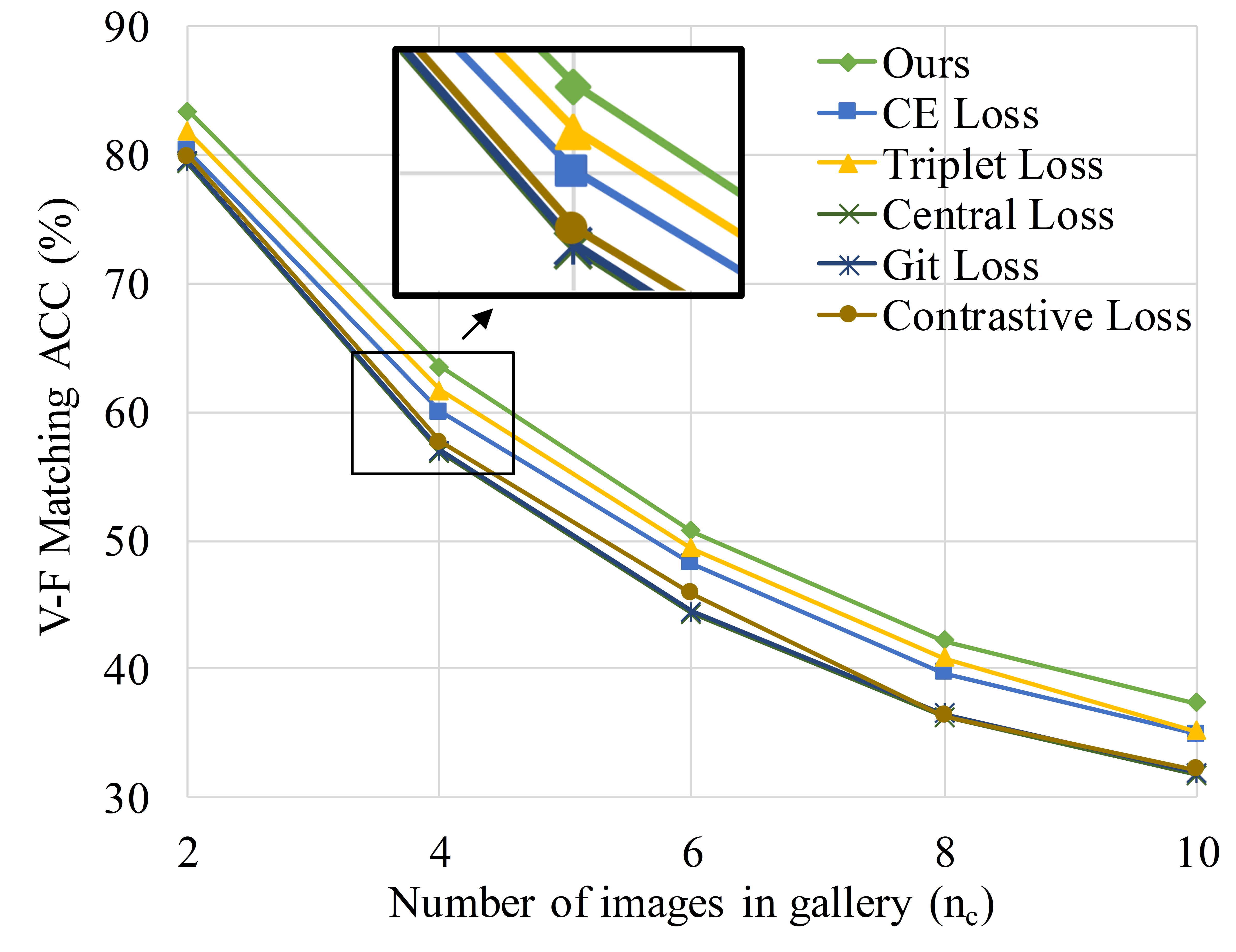}
%
%
     \includegraphics[width=0.24\textwidth]{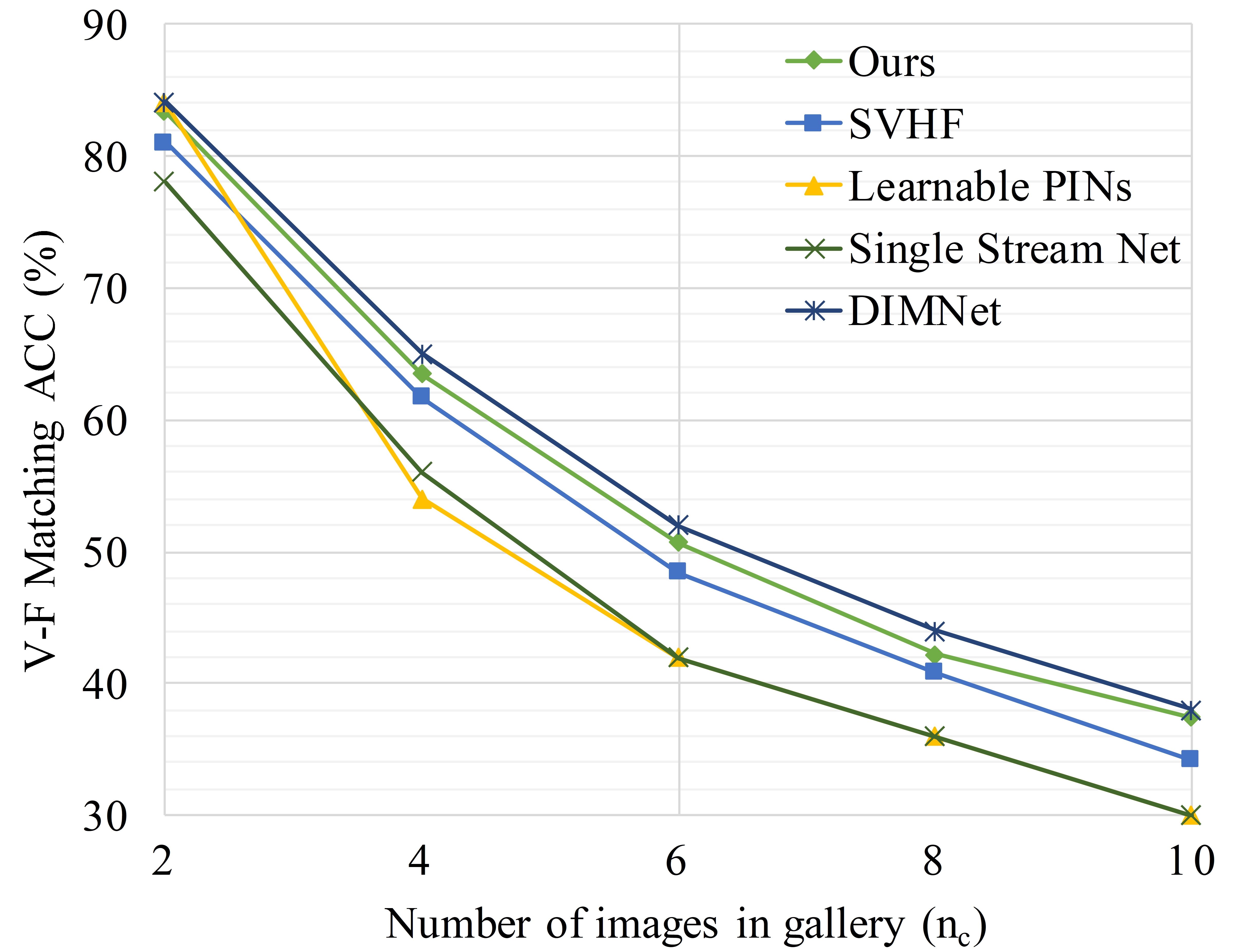}
   
  \caption{\small Cross-modal matching results: (left) FOP vs other losses used in F-V methods. (right) Our method vs state-of-the-art methods.}
\label{fig:cross_modal_matching_results}

\end{figure}

%
%
We then validate the effectiveness of our (joint) loss formulation by examining the effect of Gender (G), Nationality (N), Age (A) and its combination (GNA) separately, which influence both face and voice verification (Table~\ref{tab:results-demographic}). It achieves consistently better performance on  G, N, A and the combination (GNA) in both \textit{seen-heard} and \textit{unseen-unheard} configurations than other loss formulations.  Furthermore, we compare our (joint) loss formulation against aforementioned loss functions on a cross-modal matching task, $1:n_c$ with $n_c=2,4,6,8,10$ in Fig.~\ref{fig:cross_modal_matching_results} (left). We see that it outperforms the counterpart loss formulations for all values of $n_c$.


\begin{SCtable}[][t]
\centering
\resizebox{0.7\linewidth}{!}{
\begin{tabular}{|lcc|cc|c|}
\hline
Method  & EER & AUC & EER & AUC\\
\hline
 & \multicolumn{2}{c|}{Seen-Heard} & \multicolumn{2}{c|}{Unseen-Unheard}\\
\hline\hline
CE Loss                                                & 21.8 & 86.6  & 26.8 & 81.7\\
Center Loss~\cite{wen2016discriminative,nawaz2019deep} & 19.8 & 88.6  & 29.7 & 77.5\\
Git Loss~\cite{calefati2018git}                        & 19.6 & 88.9  & 29.5 & 77.8\\
Contrastive Loss~\cite{nagrani2018learnable}           & 23.4 & 84.7  & 29.1 & 79.5\\
Triplet Loss~\cite{schroff2015facenet}                 & 20.7 & 88.0   & 27.1 & 81.4\\ 
Ours                                                   & \textbf{19.3 }& \textbf{89.3}  & \textbf{24.9} & \textbf{83.5}\\
\hline
\end{tabular}
}
\hspace{0.05em}
\caption{\small Cross-modal verification results for our (joint) loss and other losses.}
\label{tab:result-base}
\end{SCtable}

\begin{SCtable}
\centering
\resizebox{0.60\linewidth}{!}{
\begin{tabular}{|lcc|cc|c|}
\hline
Method  & Empirical & Theoretical\\
\hline
 & Time (s)  & Worst Case\\

\hline\hline
CE Loss   & .02  & $\mathcal{O} (n)$ \\
Center Loss~\cite{wen2016discriminative,nawaz2019deep}  & 6.8 & $\mathcal{O} (n+\frac{n^2}{B})$\\
Git Loss~\cite{calefati2018git}                        &  6.2 & $\mathcal{O} (n+\frac{n^2}{B})$\\
Contrastive Loss~\cite{nagrani2018learnable}           & 568.2  & $\mathcal{O} (n^2)$\\
Triplet Loss~\cite{schroff2015facenet}                 & 619.7 & $\mathcal{O} (n^3)$\\ 
Ours                                                   & 0.7  & $\mathcal{O} (n)$ \\
\hline
\end{tabular}
}
\hspace{0.2em}
\caption{\small Theoretical and empirical run-time training complexity of our (joint) loss and others. $n$: \# training instances in a modality, and $B$ is mini-batch size.}
\label{tab:runtime-result-base}

\end{SCtable}


%
 %
 
\begin{table}[!htp]
\centering
\resizebox{0.50\textwidth}{!}{%
\begin{tabular}{|llcccc|lcccc|}
\hline
Demographic   &Rand. & G & N & A & GNA &Rand. & G & N & A & GNA \\
\hline
 & \multicolumn{5}{c|}{Seen-Heard}  & \multicolumn{5}{c|}{Unseen-Unheard}\\
\hline
CE & 86.6 & 78.0 & 85.0 & 86.3 & 77.3 & 81.7 & 65.9 & 53.6 & 76.0 & 52.8 \\
Center & 88.6  & 79.2   & 87.0  &  88.2  & 78.1  & 77.5  & 62.4  & 51.7  &  72.5  & \textbf{54.2} \\
Git  & 88.9  & \textbf{79.7}   & 87.4  &  \textbf{88.6}  & \textbf{78.5} & 77.9  & 62.6  & 51.8  &  72.8  & \textbf{54.2}   \\
Contrastive  & 84.7  & 69.7     & 83.7     &   84.5    & 69.2   & 79.5  & 61.0     & 53.5     &   74.7    & 51.8  \\
Triplet  & 88.0  & 76.3   & 86.7  &  87.6  & 75.6 & 81.7  & 65.5  & 53.4  & 76.3   & 52.2  \\
Ours                                                   & \textbf{89.3}  & 76.7  & \textbf{87.9}   &   \textbf{88.6}   &76.6  & \textbf{83.5}  & \textbf{68.8}  & \textbf{54.9}  & \textbf{78.1}   & \textbf{54.2}  \\

\hline
\end{tabular}}
\vspace{0.05em}
\caption{\small Cross-modal biometrics results under varying demographics for \textit{seen-heard} and \textit{unseen-unheard} configurations.}
\label{tab:results-demographic} \vspace{-1.5em}
\end{table}




\noindent \textbf{Comparison with state-of-the-art.} Under  \textit{unseen-unheard} protocol, our method outperforms all competing approaches, including DIMNet~\cite{wen2018disjoint}, Learnable Pins~\cite{nagrani2018learnable}, MAV-Celeb~\cite{nawaz2021cross}, Single Stream Network~\cite{nawaz2019deep}, and under \textit{seen-heard} configuration, it achieves the second best performance (see Table~\ref{tab:sota}). For cross-modal matching task, involving $1:n_c$  matching tasks, our method outperforms~\cite{nagrani2018learnable} while achieves competitive performance against DIMNet~\cite{wen2018disjoint} (Fig.~\ref{fig:cross_modal_matching_results} (right)).


\begin{table}[!htp]
\centering
\begin{tabular}{|lcc|cc|}
\hline
Methods  & EER & AUC  & EER & AUC \\
\hline
 & \multicolumn{2}{c|}{Seen-Heard} & \multicolumn{2}{c|}{Unseen-Unheard}\\
\hline\hline
DIMNet~\cite{wen2018disjoint}                          & -    & -    & 24.9  & -      \\
Learnable Pins~\cite{nagrani2018learnable}             & 21.4 & 87.0 & 29.6  & 78.5       \\
MAV-Celeb~\cite{nawaz2021cross}                        & -    & -    & 29.0  & 78.9       \\
Single Stream Network~\cite{nawaz2019deep}             & \textbf{17.2} & \textbf{91.1} &  29.5 & 78.8    \\
Ours                                                   & 19.3 & 89.3  & \textbf{24.9} & \textbf{83.5}    \\

\hline
\end{tabular}

\vspace{0.4em}
\caption{\small Cross-modal verification results of our method and existing state-of-the-art methods.}
\label{tab:sota}
\end{table}

\begin{figure}[!htp]
\begin{center}
\includegraphics[scale=0.40]{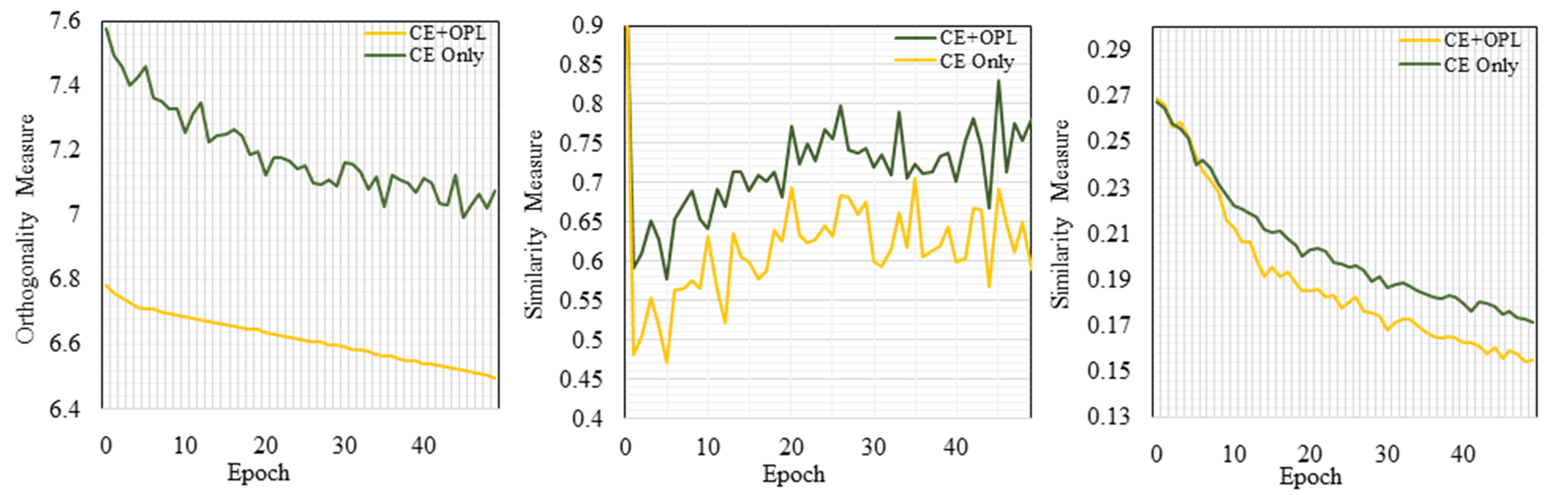}
\end{center} \vspace{-1em}
   \caption{\small (a) Feature Orthogonality (↓) (b) Similarity of same class features (↑) (c) Similarity of different class features (↓).}
   \vspace{-2em}
\label{fig:decompose}
\end{figure}

\noindent \textbf{Ablation study and analysis.} Table~\ref{tab:fusion-abl} reveals that our method's performance is mostly robust to the choice of $\alpha$, which is a hyperparameter to balance the contribution of softmax CE and orthogonal constraints based loss in our joint formulation (Eq.~\ref{eq:floss}). We also show that on replacing gated multimodal fusion with a much simpler linear fusion, the performance of our method significantly drops (Table~\ref{tab:linear-vs-gated}). Finally, in Fig.~\ref{fig:decompose}, we find that the proposed OC with CE loss, in comparison to CE loss alone, enhances (overall) feature discrminability with orthoganlity constraints, and enforces stronger intra-identity compactness and inter-identity separation in the joint F-V embedding space. 



\begin{SCtable}[][!htp]
\centering
\resizebox{0.6\linewidth}{!}{
\begin{tabular}{|l|c|c|c|c|c|c|}
\hline
 \multicolumn{7}{|c|}{$ \mathcal{L}_{CE} + \alpha \mathcal{L}_{OC}$} \\
\hline
 $\alpha$ & 0.0 & 0.1 & 0.5 & 1.0 & 2.0 & 5.0 \\
\hline\hline 
EER  & 26.8 & 26.1 & 25.8 & \textbf{24.9} & 25.9 &26.0 \\
AUC  & 81.7 & 82.4 & 82.8 & \textbf{83.5} & 82.7 & 82.6 \\
\hline
\end{tabular}
}
\hspace{0.08em}
\caption{\small Cross-modal verification results when varying $\alpha$. }
\vspace{-3em}
\label{tab:fusion-abl}

\end{SCtable}

\begin{SCtable}[][!htp]
\centering
\resizebox{0.5\linewidth}{!}{
\begin{tabular}{|l|c|c|}
\hline
Fusion Strategy & EER & AUC \\
\hline\hline
Linear Fusion           & 25.6    & 82.7 \\
Gated Fusion & \textbf{24.9}    & \textbf{83.5 } \\
\hline
\end{tabular}
}
\hspace{0.05em}
\caption{\small Cross-modal verification results with linear and gated fusion strategies.}
\vspace{-2em}
\label{tab:linear-vs-gated}
\end{SCtable}


\section{Conclusion}
We presented a light-weight module (FOP) for F-V association task. It harnesses the best in both face and voice modalities through attention-based fusion and clusters the fused embeddings based on their identity-labels via orthogonality constraints. We instantiated this module in a two-stream pipeline, used for extracting face and voice embeddings, and the resulting overall framework is evaluated on a large-scale VoxCeleb dataset for F-V matching and verification tasks. Our method performs favourably against the existing state-of-the-art methods and proposed FOP outperforms competitors both in accuracy and efficiency.


\bibliographystyle{IEEEbib}
\bibliography{IEEEbib}

\begin{thebibliography}{10}

\bibitem{kamachi2003putting}
Miyuki Kamachi, Harold Hill, Karen Lander, and Eric Vatikiotis-Bateson,
\newblock ``Putting the face to the voice': Matching identity across
  modality,''
\newblock {\em Current Biology}, vol. 13, no. 19, pp. 1709--1714, 2003.

\bibitem{belin2004thinking}
Pascal Belin, Shirley Fecteau, and Catherine Bedard,
\newblock ``Thinking the voice: neural correlates of voice perception,''
\newblock {\em Trends in cognitive sciences}, vol. 8, no. 3, pp. 129--135,
  2004.

\bibitem{nagrani2018seeing}
Arsha Nagrani, Samuel Albanie, and Andrew Zisserman,
\newblock ``Seeing voices and hearing faces: Cross-modal biometric matching,''
\newblock in {\em Proceedings of the IEEE conference on computer vision and
  pattern recognition}, 2018, pp. 8427--8436.

\bibitem{nagrani2018learnable}
Arsha Nagrani, Samuel Albanie, and Andrew Zisserman,
\newblock ``Learnable pins: Cross-modal embeddings for person identity,''
\newblock in {\em Proceedings of the European Conference on Computer Vision
  (ECCV)}, 2018, pp. 71--88.

\bibitem{nagrani2017voxceleb}
Arsha Nagrani, Joon~Son Chung, and Andrew Zisserman,
\newblock ``Voxceleb: a large-scale speaker identification dataset,''
\newblock {\em arXiv preprint arXiv:1706.08612}, 2017.

\bibitem{horiguchi2018face}
Shota Horiguchi, Naoyuki Kanda, and Kenji Nagamatsu,
\newblock ``Face-voice matching using cross-modal embeddings,''
\newblock in {\em Proceedings of the 26th ACM international conference on
  Multimedia}, 2018, pp. 1011--1019.

\bibitem{kim2018learning}
Changil Kim, Hijung~Valentina Shin, Tae-Hyun Oh, Alexandre Kaspar, Mohamed
  Elgharib, and Wojciech Matusik,
\newblock ``On learning associations of faces and voices,''
\newblock in {\em Asian Conference on Computer Vision}. Springer, 2018, pp.
  276--292.

\bibitem{nawaz2019deep}
Shah Nawaz, Muhammad~Kamran Janjua, Ignazio Gallo, Arif Mahmood, and Alessandro
  Calefati,
\newblock ``Deep latent space learning for cross-modal mapping of audio and
  visual signals,''
\newblock in {\em 2019 Digital Image Computing: Techniques and Applications
  (DICTA)}. IEEE, 2019, pp. 1--7.

\bibitem{wen2021seeking}
Peisong Wen, Qianqian Xu, Yangbangyan Jiang, Zhiyong Yang, Yuan He, and
  Qingming Huang,
\newblock ``Seeking the shape of sound: An adaptive framework for learning
  voice-face association,''
\newblock {\em arXiv preprint arXiv:2103.07293}, 2021.

\bibitem{wen2018disjoint}
Yandong Wen, Mahmoud~Al Ismail, Weiyang Liu, Bhiksha Raj, and Rita Singh,
\newblock ``Disjoint mapping network for cross-modal matching of voices and
  faces,''
\newblock in {\em 7th International Conference on Learning Representations,
  {ICLR} 2019, USA, May 6-9, 2019}, 2019.

\bibitem{nawaz2021cross}
Shah Nawaz, Muhammad~Saad Saeed, Pietro Morerio, Arif Mahmood, Ignazio Gallo,
  Muhammad~Haroon Yousaf, and Alessio Del~Bue,
\newblock ``Cross-modal speaker verification and recognition: A multilingual
  perspective,''
\newblock in {\em Proceedings of the IEEE/CVF Conference on Computer Vision and
  Pattern Recognition}, 2021, pp. 1682--1691.

\bibitem{wen2016discriminative}
Yandong Wen, Kaipeng Zhang, Zhifeng Li, and Yu~Qiao,
\newblock ``A discriminative feature learning approach for deep face
  recognition,''
\newblock in {\em European conference on computer vision}. Springer, 2016, pp.
  499--515.

\bibitem{arevalo2017gated}
John Arevalo, Thamar Solorio, Manuel Montes-y G{\'o}mez, and Fabio~A
  Gonz{\'a}lez,
\newblock ``Gated multimodal units for information fusion,''
\newblock {\em arXiv preprint arXiv:1702.01992}, 2017.

\bibitem{chen2020multi}
Zhengyang Chen, Shuai Wang, and Yanmin Qian,
\newblock ``Multi-modality matters: A performance leap on voxceleb,''
\newblock {\em Proc. Interspeech 2020}, pp. 2252--2256, 2020.

\bibitem{deng2019arcface}
Jiankang Deng, Jia Guo, Niannan Xue, and Stefanos Zafeiriou,
\newblock ``Arcface: Additive angular margin loss for deep face recognition,''
\newblock in {\em Proceedings of the IEEE/CVF Conference on Computer Vision and
  Pattern Recognition}, 2019, pp. 4690--4699.

\bibitem{hayat2019gaussian}
Munawar Hayat, Salman Khan, Syed~Waqas Zamir, Jianbing Shen, and Ling Shao,
\newblock ``Gaussian affinity for max-margin class imbalanced learning,''
\newblock in {\em Proceedings of the IEEE/CVF International Conference on
  Computer Vision}, 2019, pp. 6469--6479.

\bibitem{calefati2018git}
Alessandro Calefati, Muhammad~Kamran Janjua, Shah Nawaz, and Ignazio Gallo,
\newblock ``Git loss for deep face recognition,''
\newblock in {\em Proceedings of the British Machine Vision Conference
  ({BMVC})}, 2018.

\bibitem{ranasinghe2021orthogonal}
Kanchana Ranasinghe, Muzammal Naseer, Munawar Hayat, Salman Khan, and
  Fahad~Shahbaz Khan,
\newblock ``Orthogonal projection loss,''
\newblock {\em arXiv preprint arXiv:2103.14021}, 2021.

\bibitem{schroff2015facenet}
Florian Schroff, Dmitry Kalenichenko, and James Philbin,
\newblock ``Facenet: A unified embedding for face recognition and clustering,''
\newblock in {\em Proceedings of the IEEE conference on computer vision and
  pattern recognition}, 2015, pp. 815--823.

\bibitem{parkhi2015deep}
Omkar~M Parkhi, Andrea Vedaldi, and Andrew Zisserman,
\newblock ``Deep face recognition,''
\newblock 2015.

\bibitem{xie2019utterance}
Weidi Xie, Arsha Nagrani, Joon~Son Chung, and Andrew Zisserman,
\newblock ``Utterance-level aggregation for speaker recognition in the wild,''
\newblock in {\em ICASSP 2019-2019 IEEE International Conference on Acoustics,
  Speech and Signal Processing (ICASSP)}. IEEE, 2019, pp. 5791--5795.

\end{thebibliography}

\end{document}